\documentclass[10pt,twocolumn,letterpaper]{article}

\usepackage{cvpr}              % To produce the CAMERA-READY version
\definecolor{cvprblue}{rgb}{0.21,0.49,0.74}
\usepackage[pagebackref,breaklinks,colorlinks,allcolors=cvprblue]{hyperref}
\usepackage{afterpage}
\usepackage{fontawesome}
\usepackage{multirow}
\usepackage{xspace}
\usepackage{colortbl}
\usepackage{xcolor}
\usepackage[accsupp]{axessibility}  % Improves PDF readability for those with disabilities.
\def\paperID{17} % *** Enter the Paper ID here
\def\confName{CVPR}
\def\confYear{2026}

\title{FALO: \underline{F}ast and \underline{A}ccurate \underline{L}iDAR 3D \underline{O}bject Detection \\on Resource-Constrained Devices\vspace{-4pt}}

\author{
Shizhong Han$^{1}$~~~
Hsin-Pai Cheng$^{1}$~~~
Hong Cai$^{1}$~~~
Jihad Masri$^{2}$~~~
Soyeb Nagori$^{2}$~~~
Fatih Porikli$^{1}$~~~ 
\smallskip
\\
{$^{1}$Qualcomm AI Research\thanks{Qualcomm AI Research is an initiative of Qualcomm Technologies, Inc.}\quad$^{2}$Qualcomm Technologies, Inc.}
\\[-0pt]
\smallskip
{\tt\small\{shizhan, hsinpaic, hongcai, jmasri, soyeb, fporikli\}@qti.qualcomm.com}
\\[-6pt]
}

\newcommand{\ours}{{FALO}\xspace}
\newcommand{\ourop}{{ConvDotMix}\xspace}
\newcommand{\ccb}{\cellcolor{blue!10}}

\begin{document}
\maketitle
\afterpage{
\begin{figure}[t]
\vspace{-5pt}
  \centering
  % \fbox{\rule{0pt}{2in} \rule{0.9\linewidth}{0pt}}
  % \fbox{\rule{0pt}{2in} \rule{0.9\linewidth}{0pt}}
\includegraphics[width=0.95\columnwidth]{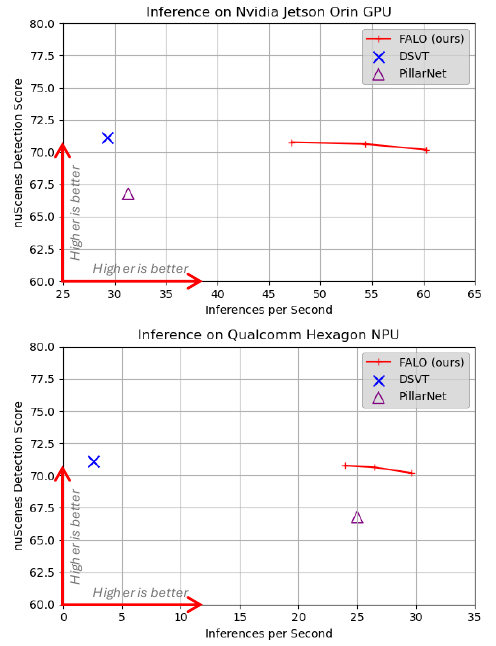}
\vspace{-8pt}
   \caption{Comparison of 3D object detection accuracy (nuScenes Detection Score on nuScenes validation) vs. inference speed (inferences per second) on two embedded platforms: Nvidia Jetson AGX Orin GPU (left) and Qualcomm\textsuperscript{\scalebox{.6}{\faRegistered}} Hexagon$^{\scalebox{.4}{\text{TM}}}$ NPU which is integrated within the Snapdragon\textsuperscript{\scalebox{.6}{\faRegistered}} 8 Gen 3 processor (right). For feasible deployment on device, we select models that do not use sparse convolutions. Specifically, DSVT~\cite{wang2023dsvt} is the only latest SOTA method that does not use sparse convolutions. Additionally, we use a modified version of PillarNet~\cite{shi2022pillarnet}, which replaces sparse convolutions with dense ones and has similar accuracy as the original model.}
   \label{fig:accuracy_latency}
   \vspace{-10pt}
\end{figure} }

\begin{abstract}
Existing LiDAR 3D object detection methods predominantly rely on sparse convolutions and/or transformers, which can be challenging to run on resource-constrained edge devices, due to irregular memory access patterns and high computational costs. 
In this paper, we propose \ours, a hardware-friendly approach to LiDAR 3D detection, which offers both state-of-the-art (SOTA) detection accuracy and fast inference speed. More specifically, given the 3D point cloud and after voxelization, \ours first arranges sparse 3D voxels into a 1D sequence based on their coordinates and proximity. The sequence is then processed by our proposed \ourop blocks, consisting of large-kernel convolutions, Hadamard products, and linear layers. \ourop provides sufficient mixing capability in both spatial and embedding dimensions, and introduces higher-order nonlinear interaction among spatial features. Furthermore, when going through the \ourop layers, we introduce implicit grouping, which balances the tensor dimensions for more efficient inference and takes into account the growing receptive field.  
All these operations are friendly to run on resource-constrained platforms and \ours can readily deploy on compact, embedded devices.
Our extensive evaluation on LiDAR 3D detection benchmarks such as nuScenes and Waymo shows that \ours achieves competitive performance. Meanwhile, \ours is 1.6$\sim$9.8$\times$ faster than the latest SOTA on NVIDIA Graphics Processing Unit (GPU) and Qualcomm Neural Processing Unit (NPU).\footnote{Snapdragon and Qualcomm branded products are products of Qualcomm Technologies, Inc. and/or its subsidiaries. Qualcomm patented technologies are licensed by Qualcomm Incorporated.}   
\vspace{-5pt}
% 3D perception base on point cloud is essential for autonomous driving, robotics, and augmented reality. However, efficient models for on-device execution are less explored compared with improving the detection accuracy. In this work, we proposed \ours, an efficient and hardware-friendly model for 3D perception on device with 3D voxel input. \ours strikes an excellent balance between computational efficiency and detection accuracy. Specifically, the 3D voxels are serialized into an ordered 1D sequence, followed by a series of proposed ConvDotMix blocks, which include multiple streams of convolutions with larget kernel size. The element-wise dot production is conducted among the outputs of the different streams. The use of only convolution and mlp operations ensures that \ours has reduced computation and faster running speed. Extensive evaluation on nuScenes and Waymo datasets demonstrate that our method achieves comparable detection accuracy to state-of-art methods while improving significant on device inference speed on both GPU and NPU.    
\end{abstract}    
% \blfootnote{${*}$Qualcomm AI Research is  initiative of Qualcomm Technologies, Inc.}
\section{Introduction}
\label{sec:intro}

Perceiving objects based on 3D LiDAR point cloud is critical to various applications, such as autonomous driving, robotics, and AR/VR. Unlike cameras, LiDAR sensors directly measure distances, and are robust to varying illumination and weather conditions, providing more reliability and safety to the autonomous system. Moreover, in recent years, LiDAR sensors have become more affordable, making it more economical to use them in production systems.

Existing methods for LiDAR-based 3D object detection predominantly utilize 3D sparse convolutions to process the 3D voxels~\cite{zhou2018voxelnet,deng2021voxel, dong2022mssvt, guan2022m3detr, liu2020tanet, shi2020points, wang2022cagroup3d, yan2018second, yin2021center, zhang2024safdnet}. While sparse convolution is more computationally and memory efficient compared to regular dense convolution, it incurs irregular data and memory access patterns due to the inherent sparsity of the 3D point cloud. This poses a challenge for compact, memory-constrained embedded devices. Furthermore, the use of sparse convolutions complicates quantization and deployment of the model via TensorRT~\cite{zhou2023fastpillars}.

Another line of work leverages pillar-based representation to project 3D voxels to a 2D plane~\cite{lang2019pointpillars}. This significantly improves the computation efficiency thanks to the reduced number of voxels. However, their detection accuracy lags behind models that directly process 3D voxels, and most recent works still require 2D sparse convolutions~\cite{li2023pillarnext,shi2022pillarnet}.

More recently, researchers have started to explore transformers for LiDAR 3D object detection~\cite{wang2023dsvt, liu2023flatformer}. For instance, in DSVT~\cite{wang2023dsvt}, the 3D voxels are treated as tokens and divided into small groups, with self-attention performed within each group. By doing this, DSVT achieves state-of-the-art performance, without using any sparse convolutions. However, transformers incur high computational and memory costs, both of which are quadratic w.r.t. the number of input tokens, \ie, non-empty voxels in this case. 
Recent works such as LION~\cite{liu2024lion} and VoxelMamba~\cite{zhang2024voxel} explore replacing transformers with Recurrent Neural Networks (RNNs) or State-Space Models (SSMs). While these models are originally designed for auto-regressive tasks, they require sequential processing of voxel tokens, which introduces inefficiencies. Moreover, they still depend on sparse convolutions for spatial reasoning. PTv3~\cite{wu2024ptv3} also aims to improve efficiency by serializing 3D data into 1D and applying Point Transformers. However, it still cannot fully eliminate the reliance on 3D sparse convolution, which is used in the xCPE layer.

In this paper, we propose a fast and accurate approach to LiDAR 3D object detection, \ours, which not only runs on embedded computation units with low latencies but also maintains competitive detection accuracy.  Figure ~\ref{fig:accuracy_latency} shows detection accuracy vs. inference speed comparisons between \ours and the latest state of the art (SOTA), where the inference speeds are measured on Nvidia Jetson Orin GPU and Hexagon NPU (included in the Snapdragon 8 Gen 3 processor), respectively, which are common choices of computation platform for autonomous systems.  
Note that except for DSVT~\cite{wang2023dsvt}, existing SOTA models cannot even be deployed due to the use of sparse convolutions. Figure~\ref{fig:accuracy_latency} demonstrates that the proposed \ours has the highest inference speed on resource-constrained computation platforms while keeping comparable detection accuracy compared with SOTA methods. 

At the core of \ours is an efficient, systematic design for processing voxels. More specifically, we first arrange the 3D voxels into a 1D sequence, based on their coordinates and proximity. Next, we utilize large-kernel convolutions and linear layers to perform spatial and channel feature mixing, respectively. In addition, we leverage Hadamard product to create higher-order nonlinear interactions among the voxel features. As shown in recent studies on efficient image backbones, such as Conv2Former~\cite{hou2024conv2former} and PADRe~\cite{letourneau2024padre}, high-order nonlinear processing is crucial for capable vision networks in lieu of the costly self-attention operations. 
Note that these operations incur only linear computation and memory costs w.r.t. the number of non-empty voxels. Moreover, we maintain a dense representation throughout the network and do not require sparse convolutions, which facilitates more efficient memory access.
We pack this series of operations in a layer, which we refer to as \ourop, and repeat the \ourop layer multiple times to process the voxel features, before converting them to Bird's Eye View (BEV).

Together with \ourop, we introduce implicit grouping to the 1D sequence that better balances the tensor dimensions. This approach improves inference speed by creating tensors with more balanced dimension sizes. We also use variable, increasing implicit group sizes through the \ourop layers, to allow for voxel interaction in wide spatial ranges as the receptive field grows larger.

In addition to not using sparse convolutions, we eliminate other cumbersome and hardware-unfriendly designs in latest SOTA methods~\cite{wang2023dsvt, liu2024lion}, \eg, window shifting. At each inference, we only need to perform 1D serialization of the voxels once, unlike existing work~\cite{wang2023dsvt, zhang2024voxel}; this significantly reduces costly data movement. 
Finally, all our operations are amenable to parallelization. In addition to these computational advantages, \ours achieves state-of-the-art detection performance, as we will see in the paper.

\begin{figure*}[t]
  \centering
   \includegraphics[width=0.85\linewidth]{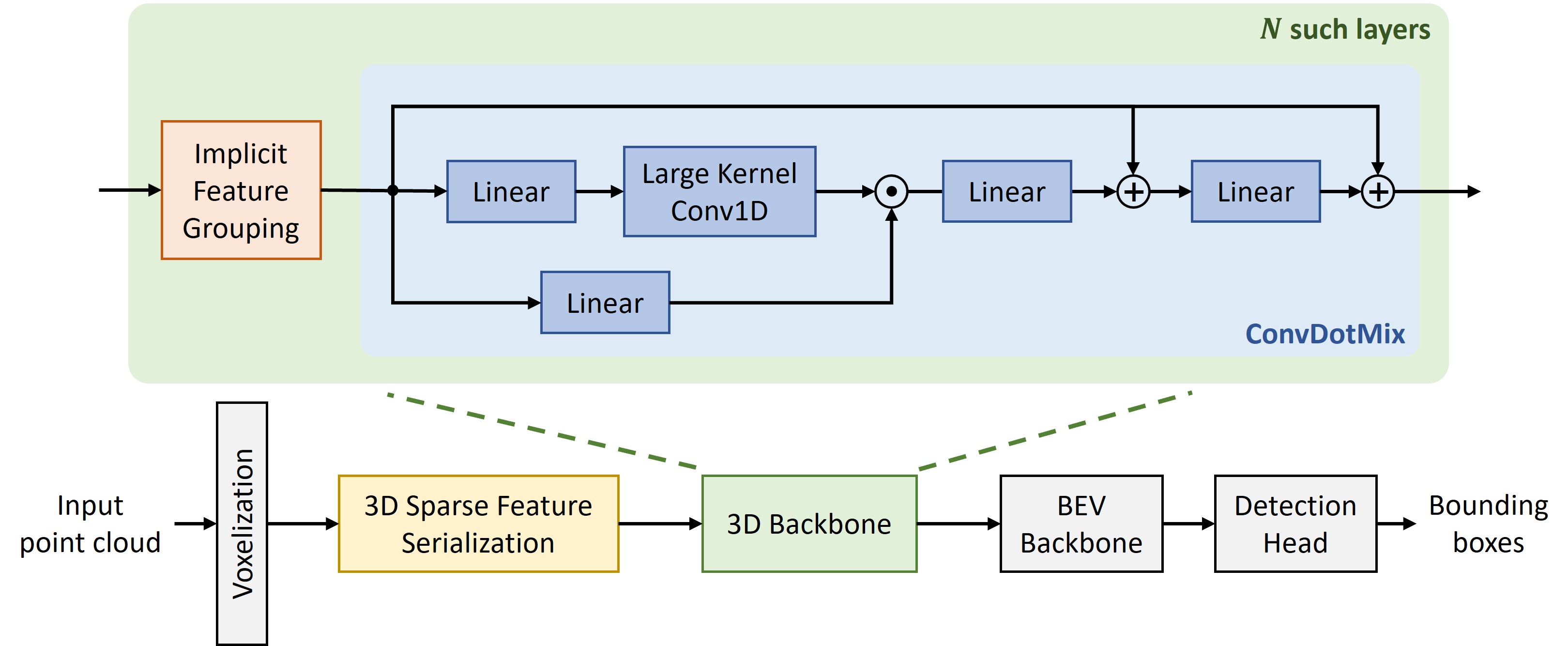}
   \vspace{-3pt}
   \caption{\small Overview of our proposed \ours approach. Given the input 3D point cloud and after voxelization, we perform 3D sparse feature serialization to arrange the non-empty 3D voxel features into a dense 1D sequence of tokens. The 1D sequence then goes into our 3D backbone, which consists of implicit feature grouping and \ours layers. Implicit grouping reshapes variables from the token dimension to the batch dimension, creating more balanced sizes across the two dimensions. \ourop operates on the tokens, mixing them spatially and channel-wise, as well as enabling higher-order nonlinear interaction between the features. $\odot$ means Hadamard product (i.e., element-wise multiplication). }
   
   % sorted to an ordered 1D sequence. Our proposed 3D sparse backbone with ConvDotMix block can efficiently extract the 3D sparse feature. The ConvDotMix block just consists of the hardware friendly operations including MLP, conv1d, and Hadamard products. The 2D backbone is proposed to scale up for detection accraucy improvement. Conv1d-k means conv1d with kernel size k. }
   \label{fig:overview}
\end{figure*}

Our main contributions are summarized as follows:

\begin{itemize}
  \item We propose \ours, an efficient, systematic approach for fast and accurate LiDAR 3D object detection on resource-constrained computation platforms. 
  \item More specifically, we propose \ourop operations to process the voxel features, which consists of large-kernel convolutions, linear layers, and Hadamard products. \ourop is computationally efficient and provides sufficient spatial, channel-wise, and higher-order nonlinear mixing capabilities to process the features.
  \item We introduce additional efficient designs that make inference faster on hardware, \eg, only performing voxel serialization once, implicit voxel grouping. Also, we eliminate deployment-unfriendly and/or costly operations used in existing SOTA, such as sparse convolution, self-attention, window shifting, and re-ordering the voxels.
  \item \ours achieves efficient inference speed on embedded computation platforms, including the Nvidia Jetson Orin GPU and Qualcomm Hexagon NPU. \ours is 1.6$\sim$9.8$\times$ faster than the latest SOTA on NVIDIA Graphics Processing Unit and Qualcomm Neural Processing Unit.
\end{itemize}

\section{Related Work}
\label{sec:related}
% \subsection{3D Object Detection in Point Cloud}
In 3D perception, two primary methodologies have emerged:  \textbf{point-based} methods and  \textbf{voxel-based} methods. Point-based methods~\cite{chen2019fast,chen2022sasa,cheng2021back,he2020structure,li2023pillarnext, pan20213d,qi2019deep,qi2018frustum,dbqssd,yang20203dssd,yang2019std,zhang2022not}, originating from the PointNet series~\cite{qi2017pointnet,qi2017pointnet++}, extract features directly from raw point clouds. Despite their performance in perception tasks, these methods are hindered by the time-consuming processes of point sampling, which incurs irregular memory access, and neighbor searching, which requires expensive computation. 

As a result, voxel-based methods~\cite{deng2021voxel,dong2022mssvt,guan2022m3detr, liu2020tanet,shi2020points,wang2022cagroup3d,yan2018second,yin2021center,zhang2024safdnet, chen2023voxelnext,zhang2024hednet} have gained prominence in recent years. These methods convert the input point clouds into a structured grid of 3D cells, known as 3D voxels, in which the grids are sparsely occupied. Instead of using the regular, dense 3D convolution, these methods employ 3D sub-manifold sparse convolution, which allows them to significantly reduce memory consumption and computational costs while improving detection accuracy. However, 3D sparse convolution requires customized CUDA operations, involves complex data structures, and incurs irregular memory accesses due to the random sparsity patterns in the data, making it expensive to run on hardware. Specifically, on resource-constrained embedded platforms, where computation capabilities and/or memory bandwidth are limited, it is very challenging, if not infeasible, to run 3D sparse convolutions. 

A special case of the voxel-based methods is the \textbf{pillar-based} approach, which combines 3D voxels at the same vertical location into a ``pillar". Pillar-based methods are significantly more efficient since there are much fewer grids to process after projecting the 3D points/voxels onto the bird's eye view~\cite{lang2019pointpillars, zhou2023fastpillars,mao2024pillarnest,li2023pillarnext,shi2022pillarnet,yin2021center}. However, pillar-based methods typically have lower accuracy when compared to voxel-based ones due to their reduced 3D representation capacity, and most of them still require 2D sparse convolutions ~\cite{li2023pillarnext,shi2022pillarnet, yin2021center}. 

Several works have explored \textbf{hybrid} approaches, leveraging the strengths of both point-based and voxel-based techniques~\cite{shi2020pv,shi2023pv,yang2023pvt}. For instance, PV-RCNN~\cite{shi2020pv} and its variants~\cite{shi2023pv,yang2023pvt} integrate point-based and voxel-based features to improve detection accuracy. While these methods aim to balance computational efficiency and feature representation capability, they often face challenges in achieving real-time performance due to the complexity of integrating multiple feature extraction schemes. 

More recently, researchers have started to explore processing the voxels in a tokenized way~\cite{wang2023dsvt,zhang2024voxel,liu2024lion}. More specifically, these approaches first serialize the 3D sparse voxels into a 1D dense sequence of tokens, which is then processed using transformers, Recurrent Neural Networks (RNNs), or State-Space Models (SSMs). All of these methods, however, heavily rely on using different orderings of the voxels. For instance, DSVT~\cite{wang2023dsvt} and LION~\cite{liu2024lion} use alternating x-order and y-order. VoxelMamba~\cite{zhang2024voxel} requires multi-scale orderings. Re-ordering the voxels several times during inference requires considerable data movement. This incurs expensive memory accesses, especially on compact devices with limited memory bandwidth. Moreover, transformers are computationally expensive with quadratic complexity w.r.t. input size, and latest SSMs, like Mamba~\cite{gu2023mamba} require customized CUDA operations, making them infeasible for non-CUDA devices.

\section{Proposed Approach: \ours}

In this section, we present our proposed fast and accurate LiDAR 3D object detection approach for edge devices, \ours. We describe our designs that enable \ours, including the serialization of 3D sparse voxels into a dense, 1D sequence of tokens (Section~\ref{sec:3Dto1D}), hardware-friendly implicit voxel grouping (Section~\ref{sec:grouping}), and the \ourop operations (Section~\ref{sec:3dbackbone}).

\subsection{System Overview}
An overview of our proposed approach is shown in Figure~\ref{fig:overview}. In the LiDAR 3D detection pipeline, the input 3D point cloud is typically first voxelized, \ie, converted into regular 3D grids that are sparsely occupied. This can be done with a voxel feature encoder, such as PointNet~\cite{qi2017pointnet}. Unlike existing works that directly apply 3D sparse convolution to process the voxel features, we extract the non-empty voxels and arrange them into a dense, 1D sequence of tokens. This facilitates using hardware-friendly, dense operations to process them in our devised 3D backbone, which includes implicit feature grouping and \ourop layers. After the 3D backbone, the voxel features are projected to the Bird's Eye View (BEV) space and further processed by 2D layers. Finally, a detection head consumes the BEV features and predicts the bounding boxes. We follow standard choices for the BEV backbone and detection head.

\begin{figure}[t]
  \centering
   \includegraphics[width=0.75\linewidth]{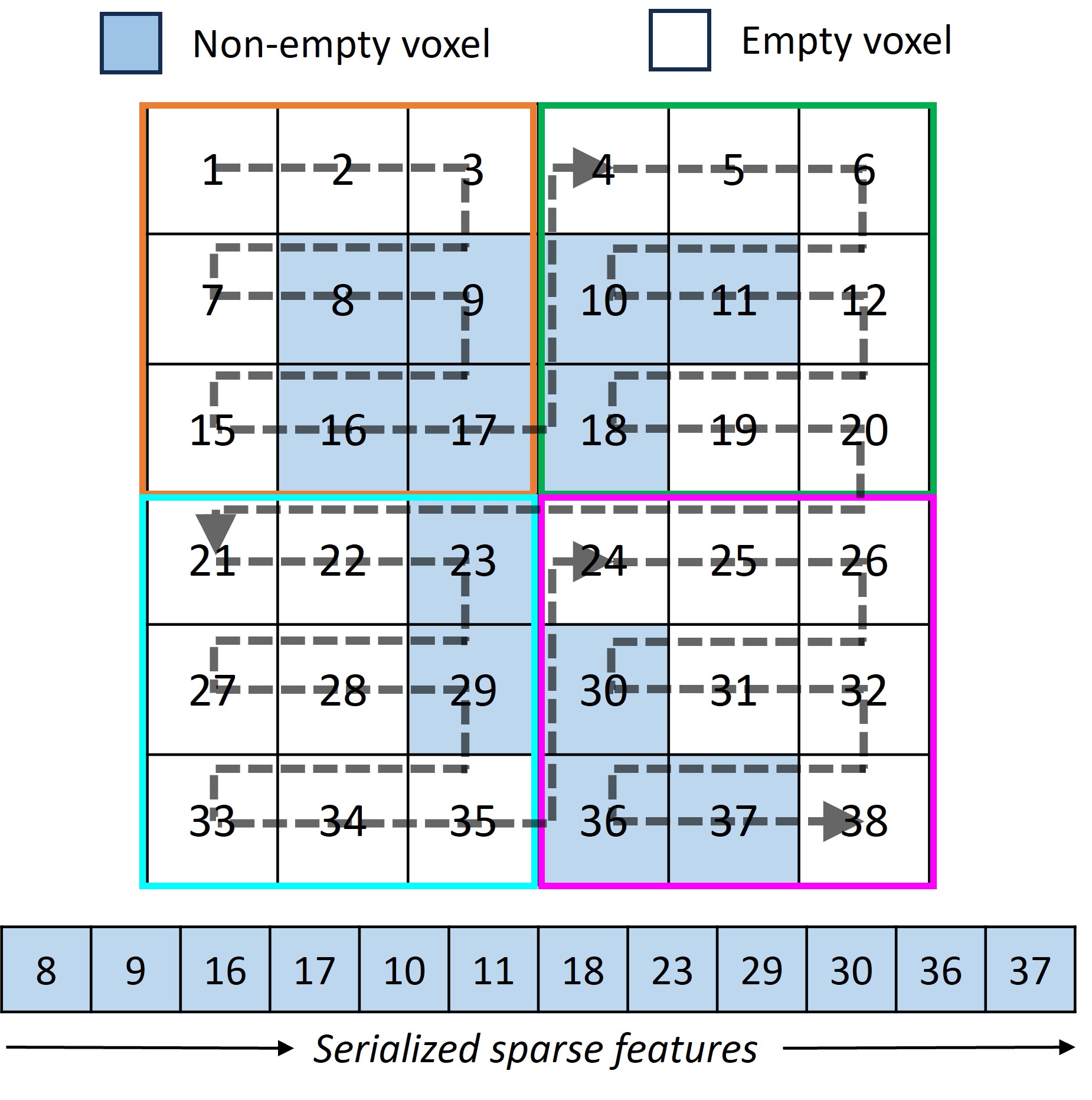}
   \vspace{-9pt}
   \caption{We divide the space into local windows (shown in different colors) to preserve locality. Within each window, we traverse and collect non‑empty voxels, forming a 1D dense sequence of tokens before moving to the next window.}
   \vspace{-0pt}
   \label{fig:serialization}
\end{figure}

%summery paragraph
\subsection{3D Sparse Feature Serialization}
\label{sec:3Dto1D}

Existing methods use expensive ways to serialize the voxel features, \eg, alternating between x- and y-orders~\cite{wang2023dsvt,liu2024lion}, using forward, backward, Z-order~\cite{wu2024ptv3}, Hilbert order~\cite{{wu2024ptv3}} and multi-scale orders~\cite{zhang2024voxel}. What is more, these methods all require re-ordering the voxels multiple times throughout the network layers. This creates costly data movements, especially for devices with limited memory bandwidth.

In \ours, we use an efficient and effective way to arrange the non-empty voxels into a 1D sequence. Figure~\ref{fig:serialization} provides an illustrative example of our approach. First, we divide the voxels into local windows (highlighted with different colors in the figure) for the locality preservation, and traverse the voxels within each window before proceeding to the next window. In this way, the non-empty voxels in the same local neighborhood reside in nearby locations in the 1D sequence, which allows their features to interact more easily via convolution operations. Within each window, it is possible to follow x- or y-order to traverse the voxels and collect the non-empty ones into the 1D sequence. Figure~\ref{fig:serialization} shows an example of using the x-order. In our experiments, we observe minimal performance difference between using x-order and y-order. 

Note that our serialization operation is only performed once after voxelization. Then, the order of the 1D sequence of voxel tokens remains the same throughout the 3D backbone. This requires minimal data movement to arrange the voxels into a 1D sequence.

\begin{table}[t!]
    \small
    \centering
    % \resizebox{0.9\linewidth}{!}{
    \begin{tabular}{c|c}
        \hline
        Input Tensor Shape & Latency on NPU (ms)  \\
        \hline
        
        $1\times6000\times128$ & 10.60 \\
         $2\times3000\times128$ & 9.53 \\
         $4\times1500\times128$ & 9.00 \\
         $10\times600\times128$ & 8.40 \\
         $20\times300\times128$ & 8.12 \\
         $60\times100\times128$ & 7.99 \\
        \hline
    \end{tabular}
    % }
    \vspace{-5pt}
    \caption{\small Latencies of inference one \ourop layer with different tensor shape. The input shape is in batch size $\times$ number of tokens $\times$ embedding dimension. Latency is measured on the Qualcomm Hexagon NPU.}
    \label{tab:dim_bal}
    \vspace{-5pt}
\end{table}

\begin{figure*}[t]
  \centering
   \includegraphics[width=0.85\linewidth]{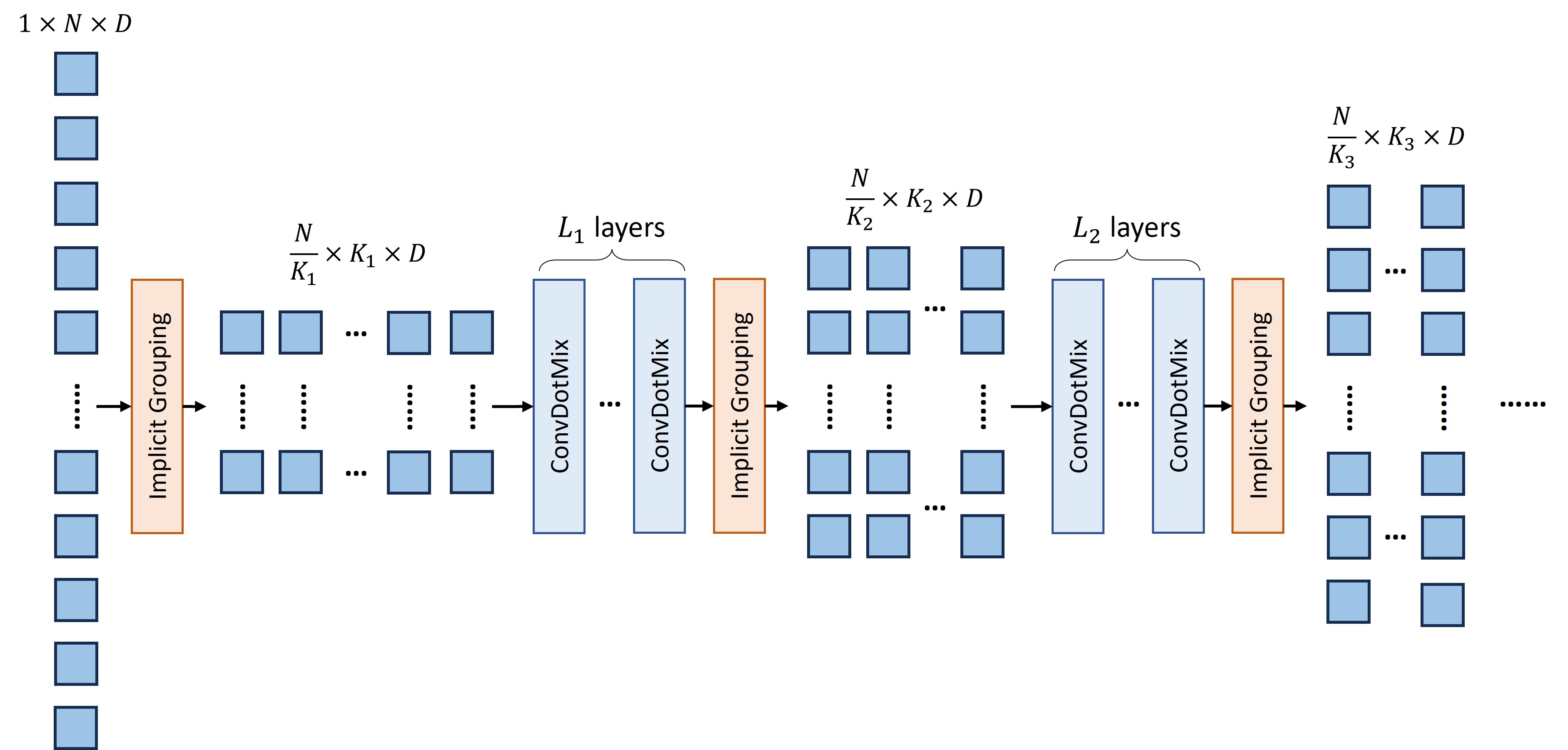}
   \caption{Implicit voxel grouping. The 1D sequence of non-empty voxel tokens with shape $1\times N\times D$ is converted to $N/K_1 \times K_1 \times D$, which implicitly creates $N/K_1$ groups of $K_1$ tokens each. As we proceed to deeper \ourop layers, we use bigger group sizes, \eg, $K_2$, $K_3$, where $K_1 < K_2 < K3$. Smaller group sizes at the beginning facilitate faster inference and more focused local learning, while larger group sizes at deeper layers work better with the larger effective receptive field.}
   \vspace{-5pt}
   \label{fig:grouping}
\end{figure*}

\subsection{Implicit Voxel Grouping} 
\label{sec:grouping}
Given that the number of non-empty voxels is usually in the order of thousands, the dense 1D sequence has a severely imbalanced shape, \eg, $1\times6000\times128$, where 1 is the batch size, 6000 is the number of non-empty voxels, and 128 is the feature embedding dimension. We observe that this is less efficient on a compact embedded device, as compared to the case where the dimension sizes are more balanced. We speculate that when the dimensions are more balanced, potential underlying parallelization may be utilized to enable faster inference.  
To make this more concrete, we profile one layer of \ourop on the Qualcomm Hexagon NPU with different input tensor shapes, the result of which is shown in Table~\ref{tab:dim_bal}. Indeed, we can see that when the sizes are more balanced across the batch and token dimensions, the inference is faster.

Motivated by this observation, we propose to reshape the 1D sequence and better balance the batch and token dimensions, which implicitly groups the voxels. For instance, when changing the shape from $1\times6000\times128$ to $10\times600\times128$, the 6000 voxels are divided into 10 groups of 600 voxels each. 
While this improves inference speed, such grouping limits the interaction among the voxels, especially when the group size is small. In order to address this, we further propose to use variable, increasing group sizes as we go to deeper \ourop layers. This is because \ourop uses convolutions to perform spatial mixing and with more \ourop layers, the effective receptive field becomes larger and creates wider interaction among the voxels. By increasing the group size in deeper layers, we mitigate its impact on the growing receptive field. 

Figure~\ref{fig:grouping} provides an illustration. In earlier layers, we use smaller group sizes, which provide more balanced sizes across the batch and token dimensions for more efficient inference. In deeper layers, we use larger group sizes to account for larger effective receptive fields. Another benefit of using increasing group sizes is that this encourages the network to first aggregate local information, before incorporating bigger spatial contexts in deeper layers. As we shall see in our experiments, increasing group sizes leads to the best performance while a decreasing order is detrimental.

% numbers from PillarNeST
\begin{table*}[t!]
    \footnotesize
    \centering
    \resizebox{0.99\linewidth}{!}{
    \begin{tabular}{c|c|c|c|c| c c c c c c c c c c}
        \hline
        Repr. & Method & Key Operators & NDS & mAP & Car & Truck & Bus & T.L. & C.V. & Ped. & M.T. & Bike & T.C. & B.R. \\
        \hline
        \multirow{9}{*}{Voxel} & CenterPoint~\cite{yin2021center} & SpConv & 64.8 & 56.4 & - & - & - & - & - & - & - & - & - & - \\
        
        & VoxelNeXt~\cite{chen2023voxelnext} & SpConv & 66.7 & 60.5 & 83.9 & 55.5 & 70.5 & 38.1 & 21.1 & 84.6 & 62.8 & 50.0 & 69.4 & 69.4  \\
        & VISTA~\cite{deng2022vista} & SpConv & 68.1 & 60.8 & 84.8 &57.2 &20.5 &67.6 &36.8 &69.0 &67.7 &50.7 &84.1 &69.7  \\
        & Uni3DETR ~\cite{wang2024uni3detr} & SpConv & 68.5 & 61.7 &-&-&-&-&-&-&-&-&-&- \\
        & TransFusion-L~\cite{bai2022transfusion} & SpConv & 70.1 & 65.5 & 86.9 & 60.8 & 73.1 & 43.4 & 25.2 & 87.5 & 72.9 & 57.3 & 77.2 & 70.3  \\

        & LargeKernel3D~\cite{chen2023largekernel3d} & SpConv & 69.1 & 63.9& 85.1& 60.1& 72.6& 41.4& 24.3& 85.6& 70.8& 59.2& 72.3& 67.7 \\

        & LinK~\cite{lu2023link} & SpConv & 69.5 & 63.6&-&-&-&-&-&-&-&-&-&-  \\
        
        & FSDv2~\cite{fan2023fsd} & SpConv & 70.4 &64.7 &84.4 &57.3 &75.9 &44.1 &28.5 &86.9 &69.5 &57.4 &72.9 &73.6  \\
        & SAFDNet\cite{zhang2024safdnet} & SpConv &71.0 &66.3 &87.6 &60.8 &78.0 &43.5 &26.6 &87.8 &75.5 &58.0 &75.0 &69.7  \\

        \hline  
        \multirow{8}{*}{Pillar} 
        & CenterPoint-Pillars~\cite{yin2021center} & SpConv & 60.2 & 50.3&-&-&-&-&-&-&-&-&-&- \\
        %& PillarNet-18~\cite{shi2022pillarnet} & SpConv & 67.4 & 59.9&-&-&-&-&-&-&-&-&-&- \\
        & PillarNet-34~\cite{shi2022pillarnet} & SpConv & 67.6 & 60.2  &-&-&-&-&-&-&-&-&-&- \\
        & FastPillars-m~\cite{zhou2023fastpillars} & Conv, Attention & 68.2 & 61.7 &-&-&-&-&-&-&-&-&-&- \\
        & PillarNeXt-B~\cite{li2023pillarnext} & SpConv & 68.8 & 62.5 &-&-&-&-&-&-&-&-&-&-  \\
        & PillarNeSt-Base~\cite{mao2024pillarnest} & Conv, MLP & 69.2 & 63.2 &-&-&-&-&-&-&-&-&-&-  \\

        % & HEDNet~\cite{zhang2024hednet} & SConv & 71.4 & 66.7  \\
  
        % LION-Mamba~\cite{liu2024lion} & SConv\&RNN & 72.1 & 68.0 & 87.9 & 64.9 & 77.6 & 44.4 & 28.5 & 89.6 & 75.6 & 59.4 & 80.8 & 71.6 \\
        % \hline
        & DSVT~\cite{wang2023dsvt} & Transformer & 71.1 & 66.4 & 87.4 & 62.6 & 75.9 & 42.1 & 25.3 & 88.2 & 74.8 & 58.7 & 77.9 & 71.0 \\
        & \ccb \ours (Ours) &\ccb Conv, MLP, HP &\ccb 70.8 &\ccb 65.7&\ccb 87.1 &\ccb 62.3 &\ccb 75.3 &\ccb 43.0 &\ccb 25.5 &\ccb 87.6 &\ccb 71.9 &\ccb 57.9 &\ccb 77.5 &\ccb 68.7 \\
        \hline
    \end{tabular}
    }
    \vspace{-5pt}
    \caption{\small Performance comparison on nuScenes validation set. HP stands for Hadamard product. T.L., C.V., Ped., M.T., T.C., and B.R. stand for trailer, construction vehicle, pedestrian, motorcycle, traffic cone, and barrier, respectively.}
    \label{tab:nuscenes_val}
\end{table*}

% numbers from PillarNeST
\begin{table*}[t!]
    \footnotesize
    \centering
    \resizebox{0.99\linewidth}{!}{
    \begin{tabular}{c|c|c|c|c| c c c c c c c c c c}
        \hline
        Repr. & Method & Key Operators & NDS & mAP & Car & Truck & Bus & T.L. & C.V. & Ped. & M.T. & Bike & T.C. & B.R.\\
        \hline
        \multirow{7}{*}{Voxel} & CenterPoint~\cite{yin2021center} & SpConv & 65.5 & 58.0 & 84.6 &51.0 &60.2 &53.2 &17.5 &83.4 &53.7 &28.7 &76.7 &70.9 \\
        
        & VoxelNeXt~\cite{chen2023voxelnext} & SpConv & 70.0 & 64.5 &84.6 &53.0 &64.7 &55.8 &28.7 &85.8 &73.2 &45.7 &79.0 &74.6 \\
        
        & TransFusion-L~\cite{bai2022transfusion} & SpConv & 70.2 &65.5 &86.2 &56.7 &66.3 &58.8 &28.2 &86.1 &68.3 &44.2 &82.0 &78.2\\

        & LargeKernel3D~\cite{chen2023largekernel3d} & SpConv & 70.5 & 65.3  &85.9 &55.3 &66.2 &60.2 &26.8 &85.6 &72.5 &46.6 &80.0 &74.3\\

        & LinK~\cite{lu2023link} & SpConv & 71.0 & 66.3 &86.1 &55.7 &65.7 &62.1 &30.9 &85.8 &73.5 &47.5 &80.4 &75.5 \\
        & FSDv2~\cite{fan2023fsd} & SpConv & 71.7 &66.2 &83.7 &51.6 &66.4 &59.1 &32.5 &87.1 &71.4 &51.7 &80.3 &78.7 \\
        & SAFDNet\cite{zhang2024safdnet} & SpConv &72.3 &68.3 &87.3 &57.3 &68.0 &63.7 &37.3 &89.0 &71.1 &44.8 &84.9 &79.5 \\

        % & SFDNet & SpConv & 71.0 & 66.3 \\

        \hline  
        \multirow{4}{*}{Pillar} & PointPillars~\cite{lang2019pointpillars} & MLP & 45.3 & 30.5 &68.4 &23.0 &28.2 &23.4 &4.1 &59.7 &27.4 &1.1& 30.8& 38.9 \\
        %& PillarNet-18~\cite{shi2022pillarnet} & SpConv & 70.8 &65.0 &87.4 &56.7 &60.9 &61.8 &30.4 &87.2 &67.4 &40.3 &82.1 &76.0 \\
        %& PillarNet-34~\cite{shi2022pillarnet} & SpConv & 71.4 &66.0 &87.6 &57.5 &63.6 &63.1 &27.9 &87.3 &70.1 &42.3 &83.3 &77.2 \\
        
        & PillarNeSt-Base~\cite{mao2024pillarnest} & Conv, MLP & 71.3 & 65.6 &87.1 &55.5 &61.6 &62.1 &31.0 &86.3 &69.4 &46.8 &80.6 &76.0 \\
        & DSVT~\cite{wang2023dsvt} & Transformer & 72.7 &68.4 &86.8 &58.4 &67.3 &63.1 &37.1 &88.0 &73.0 &47.2 &84.9 &78.4\\
        & \ccb \ours (Ours) &\ccb Conv, MLP, HP &\ccb 70.9 &\ccb 66.1 &\ccb 85.8 &\ccb 56.3 &\ccb 66.7 &\ccb 60.8 &\ccb 33.2 &\ccb 87.0 &\ccb 69.3 &\ccb 44.7 &\ccb 84.1 &\ccb 73.0\\
        \hline
    \end{tabular}
    }
    \vspace{-5pt}
    \caption{\small Performance comparison on nuScenes test set. All reported results are without applying test-time augmentation or model ensemble. HP stands for Hadamard product.}
    \label{tab:nuscenes_test}
    \vspace{-10pt}
\end{table*}

\subsection{\ourop}
\label{sec:3dbackbone}
Our devised 3D backbone contains multiple \ourop layers (in addition to the implicit voxel grouping operations). Each \ourop layer is made of large-kernel convolutions operating on the tokens (\ie, non-empty voxels), linear layers operating on the embedding dimension, as well as Hadamard product (\ie, element-wise multiplication) that creates higher-order nonlinear interaction among the voxels. More specifically, as shown in Figure~\ref{fig:overview}, the sequence of non-empty voxels (after grouping) goes through linear layers and larger-kernel convolutions. At the same time, a copy of the voxels are mixed channel-wise but without spatial mixing. These two differently mixed versions of voxels are combined via Hadamard product. In this way, the tokens can interact with each other beyond linear operations. As the effective receptive field becomes larger after several convolution layers, such nonlinear interactions can happen even between two spatially distant voxels. 
As inspired by MetaFormer~\cite{yu2022metaformer}, we further process the combined features through additional linear layers, as well as add skip connections from the input.

All operations in \ourop incur only linear computation and memory costs with respect to the number of input tokens. Our use of Hadamard product follows recent efficient alternative self‑attention designs~\cite{hou2024conv2former, letourneau2024padre}, which show such operations can replace standard self‑attention while maintaining linear complexity. Moreover, the \ourop components—convolution, MLP, and element‑wise multiplication—reduce to matrix multiplications, which are well supported and parallelizable on modern AI hardware.

\section{Experiments}
We perform extensive evaluation of both the efficiency and accuracy of our proposed \ours approach. More specifically, we measure the latency of our model and compare with the latest SOTA on both mobile Orin GPU and Hexagon NPU. We evaluate the detection accuracy on standard benchmarks for LiDAR 3D detection. Further, we conduct extensive ablation study to analyze our proposed designs.

\begin{table*}[h!]
    \footnotesize
    \centering
    \resizebox{0.99\linewidth}{!}{
    \begin{tabular}{l|c|c|c|c|c|c|c|c|c}
        \hline
        \multirow{2}{*}{Method} & \multirow{2}{*}{Operator} & \multicolumn{2}{c|}{Vehicle 3D AP/APH} & \multicolumn{2}{c|}{Pedestrian 3D AP/APH} & \multicolumn{2}{c|}{Cyclist 3D AP/APH} & mAP/mAPH & mAP/mAPH\\
        & & L1 & L2 & L1 & L2 & L1 & L2 & L1 & L2 \\
        \hline
        %SECOND~\cite{yan2018second} & SpConv & 72.3/71.7 & 63.9/63.3 & 68.7/58.2& 60.7/51.3 &60.6/59.3 &58.3/57.0 &67.2/63.1 &61.0/57.2 \\
        
        PointPillar~\cite{lang2019pointpillars} & SpConv & 72.1/71.5 &63.6/63.1 &70.6/56.7 &62.8/50.3 &64.4/62.3 &61.9/59.9 &69.0/63.5 &62.8/57.8 \\
        
        CenterPoint~\cite{yin2021center}  & SpConv & 74.2/73.6 & 66.2/65.7 & 76.6/70.5 & 68.8/63.2 &72.3/71.1 & 69.7/68.5 & 74.4/71.7& 68.2/65.8\\

        PillarNet-34~\cite{shi2022pillarnet} & SpConv & 79.1/78.6 &70.9/70.5 &80.6/74.0 &72.3/66.2 &72.3/71.2 &69.7/68.7 & 77.3/74.6&71.0/68.5 \\

        PillarNeXt~\cite{li2023pillarnext} & SpConv & 78.4/77.9 &70.3/69.8 &82.5/77.1 &74.9/69.8 &73.2/72.2 &70.6/69.6 & 78.0/75.7&71.9/69.7 \\
  
        VoxelNext~\cite{chen2023voxelnext} & SpConv & 78.2/77.7 &69.9/69.4 &81.5/76.3 &73.5/68.6 &76.1/74.9 &73.3/72.2 & 78.6/76.3 &72.2/70.1 \\
        
        Transfusion~\cite{bai2022transfusion} & SpConv & –/– &–/65.1 &–/– &–/63.7 &–/– &–/65.9 & -/- &–/64.9 \\
        FocalFormer3D~\cite{chen2023focalformer3d}  & SpConv & –/– &68.1/67.6 &–/– &72.7/66.8 &–/– &73.7/72.6 & -/- &71.5/69.0 \\
        HEDNet~\cite{zhang2024hednet}  & SpConv &81.1/80.6 &73.2/72.7 &84.4/80.0 &76.8/72.6 &78.7/77.7 &75.8/74.9 &81.4/79.4 &75.3/73.4 \\
        
        SST\_TS~\cite{fan2022embracing}  & Transformer &76.2/75.8 &68.0/67.6 &81.4/74.0 &72.8/65.9 &–/– &–/– & -/- &–/–\\
        SWFormer~\cite{sun2022swformer} & Transformer &77.8/77.3 &69.2/68.8 &80.9/72.7 &72.5/64.9 &–/– &–/– & -/- &–/– \\
        OcTr~\cite{zhou2023octr}& Transformer &78.1/77.6 &69.8/69.3 &80.8/74.4 &72.5/66.5 &72.6/71.5 &69.9/68.9 & 77.2/74.5 &70.7/68.2 \\
        DSVT-Pillar~\cite{wang2023dsvt} & Transformer &79.3/78.8 &70.9/70.5 &82.8/77.0 &75.2/69.8 &76.4/75.4 &73.6/72.7 &79.5/77.1 &73.2/71.0\\
        % DSVT-Voxel~\cite{wang2023dsvt}  & Transformer & 79.7/79.3 &71.4/71.0 &83.7/78.9 &76.1/71.5 &77.5/76.5 &74.6/73.7 &80.3/78.2 &74.0/72.1\\
        %LION~\cite{liu2024lion}  & SConv\&RNN &79.5/79.1 &71.1/70.7 &84.9/80.4 &77.5/73.2 &79.7/78.7 &76.7/75.8 &83.1/80.0 &75.1/73.2\\
        %FoxDet-Pillar (Ours) & - & CNN & 75.5/78.0 & 70.1/69.7 & 81.7/75.8 & 74.0/68.5 & 72.9/71.9 & 70.2/69.2 & - \\
        FlatFormer~\cite{liu2023flatformer} & Transformer & -/- & 69.0/68.6 & -/- & 71.5/65.3 & -/- & 68.6/67.5 & -/- & -/67.2\\
        FastPillars~\cite{zhou2023fastpillars} & Conv & -/- & 71.5/71.1 & -/- & 73.2/67.2 & -/- & 70.5/69.5 & -/- & -/69.3 \\
        PTv3~\cite{wu2024ptv3} & SpConv & -/- & 71.2/70.8 & -/- & 76.3/70.4 & -/- & 71.5/70.4 & -/- & -/70.5 \\
        \ccb \ours (Ours)  & \ccb Conv, MLP, HP & \ccb 78.7/78.3 &\ccb 70.3/69.6 &\ccb 82.6/76.9 &\ccb 75.0/69.6 &\ccb 75.4/74.4 &\ccb 75.4/72.5 &\ccb 78.9/76.5 &\ccb 73.6/70.6  \\
        \hline
        
    \end{tabular}}
    \vspace{-5pt}
    \caption{Performance comparison on the Waymo Open Dataset validation set. HP stands for Hadamard product. All experiments are evaluated with 1 input frame.}
    \vspace{-10pt}
    \label{tab:waymo}
\end{table*}

\subsection{Datasets and Evaluation Protocol}
\textbf{nuScenes dataset}~\cite{caesar2020nuscenes} is a widely recognized and challenging benchmark for outdoor 3D perception, offering a perception range of up to 50 meters. It provides comprehensive annotations for various tasks, including 3D object detection and BEV map segmentation. Each frame is annotated at a frequency of 2Hz. The dataset comprises 1,000 scenes, divided into 750 scenes for training, 150 scenes for validation, and 150 scenes for testing, resulting in a total of 40,157 annotated samples. Each sample includes data from six cameras and a 32-beam LiDAR scan. 

For 3D object detection, the dataset employs mean Average Precision (mAP) and the nuScenes Detection Score (NDS) as evaluation metrics. Note that when evaluating model performance, we do not use test-time augmentation or ensemble techniques. These techniques are not practical for real-world applications as they significantly increase the inference cost, \eg, the commonly used double-flip augmentation requires 4 inferences per input sample and quadruples the latency.

\textbf{Waymo Open Dataset} (WOD) ~\cite{sun2020scalability} is a prominent and well-known benchmark for large-scale outdoor 3D perception. It consists of 1,150 point cloud sequences, totaling over 200,000 frames. Each frame spans a substantial perception range of 150m$\times$150m. The dataset is divided into 798 scenes for training, 202 scenes for validation, and 150 scenes for testing, with each scene containing approximately 200 frames. 

For evaluation, WOD employs 3D mean Average Precision (mAP) and mean Average Precision weighted by Heading accuracy (mAPH). These metrics are further divided into two difficulty levels: L1 for objects detected with more than five points and L2 for those detected with at least one point. Similarly, we do not use any test-time augmentation or model ensemble here.

\subsection{Implementation Details}
We use 8 \ourop layers in our 3D backbone, with a kernel size 11 for the 1D convolutions and group sizes of 128, 128, 256, 256, 512, 512, 1024, and 1024. No reshaping is needed between layers that use the same group size. We use y-order to serialize the voxels, and there is minimal accuracy difference between using x-order and y-order. Due to the different LiDAR sensors, point cloud densities, and evaluation ranges, we use a pillar size of (0.3m, 0.3m, 8m) for nuScenes data and (0.32m, 0.32m, 6m) for Waymo data, following standard practice. Similar to DSVT~\cite{wang2023dsvt}, we use ResNet layers~\cite{he2016deep} for the BEV backbone. In the DSVT setup, there are 3 stages of ResNet with 2, 3, and 3 layers each. We create 2 more variants using 3-4-4 and 4-5-5 configurations, respectively. For the detection head, we use the TransFusion~\cite{bai2022transfusion} head for nuScenes and the CenterPoint~\cite{yin2021center} head for Waymo. 

% We design two different models for the evaluation datasets nuScenese and Waymo, due to the different LiDAR sensors, point cloud density, and evaluation ranges required for comparison with the SOTA methods. For the nuScenes dataset, the pillar size is set to (0.3m, 0.3m, 8m) following the DSVT~\cite{wang2023dsvt}. For the Waymo dataset, the pillar size is (0.32m, 0.32m, 6m) following the CenterPoint-Pillars~\cite{yin2021center}. 

% Across 8 blocks, we use dynamic and increased group size in sequence dimension as (128, 128, 256, 256, 512, 512, 1024, 1024). For the window partition and serialization, we mainly follow the baseline model DSVT. The different ordering has very small impact on the final detection accuracy, we just select the Y ordering serialization. 

We choose Nvidia Jetson Orin GPU and Qualcomm Hexagon NPU as the embedded edge compute platforms to measure model latency because they are widely adopted and easily accessible edge‑AI platforms, representing both GPU- and NPU‑based architectures. For running a model on device, we first export it to ONNX format, followed by W8A8 quantization. We further use TensorRT when running the model on the Orin GPU. For comparison, we also measure the inference speed of DSVT~\cite{wang2023dsvt}, which is the only available model with SOTA detection accuracy and does not require sparse convolutions.\footnote{While FastPillars~\cite{zhou2023fastpillars} and PillarNeSt~\cite{mao2024pillarnest} also claim SOTA performance without using sparse convolutions, they have not made their models publicly available.} We also compare with a modified version of PillarNet~\cite{shi2022pillarnet}, which is a representative recent pillar-based model. To enable on-device inference, we replace its sparse convolutions with dense ones and have verified that the modified version achieves similar detection accuracy as the original model via retraining.

\subsection{Detection Performance Evaluation}

% Unlike existing methods that rely on transformers and sparse convolutions, our approach utilizes a streamlined convolution-based operation for the 3D backbone, enhancing both simplicity and efficiency. We evaluate our model, FALO, on two publicly available datasets, NuScenes and Waymo, demonstrating its effectiveness across multiple metrics.

\textbf{nuScenes}. Table~\ref{tab:nuscenes_val} and Table~\ref{tab:nuscenes_test} summarize the detection performance on nuScenes validation and test sets, respectively. We see that \ours achieves similar detection performance compared to the latest SOTA, \eg, DSVT~\cite{wang2023dsvt}, SAFDNet~\cite{zhang2024safdnet}, despite not using expensive, deployment-unfriendly operations like self-attention and 3D sparse convolution. It is noteworthy that without using the 3D voxel representation and 3D sparse convolution, \ours outperforms most voxel-based methods, and additionally, \ours provides significantly better detection accuracy as compared to most pillar-based methods. While PillarNeSt-Base~\cite{mao2024pillarnest} has slightly better accuracy than \ours, it uses the heavy ConvNeXt-Base~\cite{liu2022convnet} as the BEV backbone.

\textbf{Waymo}. As summarized in Table~\ref{tab:waymo}, our proposed \ours also delivers competitive performance on the Waymo dataset. It achieves detection accuracy on par with SOTA, achieving comparable Average Precision (AP) and Average Precision with Heading (APH) scores across key classes, including vehicles, pedestrians, and cyclists. This result reinforces the generalization and robustness of our proposed efficient \ours architecture across different datasets.

\subsection{Latency Evaluation}
Figure~\ref{fig:accuracy_latency} shows the latency comparison between our proposed \ours models with DSVT~\cite{wang2023dsvt} and PillarNet~\cite{shi2022pillarnet}. It can be seen that with similar detection performance, \ours achieves significantly higher inference speeds that are 1.6$\times$$\sim$9.8$\times$ faster than DSVT on Nvidia Jetson Orin GPU and Qualcomm Hexagon NPU. Specifically, \ours offers 24$\sim$60 inferences per second on device, enabling real-time, accurate 3D LiDAR object detection operations on mobile autonomous systems, such as self-driving cars and mobile robots. Although PillarNet can also provide $\sim$30 inferences per second, its detection accuracy is considerably lower due to the less capable network architecture.

\subsection{Ablation Study}
In this part, we analyze the effects of various designs in our proposed approach. 

% conduct ablation studies to assess the impact of each component within our proposed method, including the replacement of transformers with proposed ConvDotMix in 3D backbone, as well as design contributions such as scaled-up 2D backbone, group partitioning and the impact of kernel size to the detection accuracy. These evaluations help to clarify the contribution of each element to the overall performance and efficiency of our model.

\textbf{Ordering.}
When serializing the voxels, our proposed ordering is based on local window partitions and traversing voxels in the same neighborhood first before proceeding to the next window. Here, we compare with an variant where no explicit ordering is used, \ie, the voxels are traversed based on their pre-existing order which depends on when the corresponding point is measured by the LiDAR sensor. Table~\ref{tab:ordering} shows that our proposed ordering is effective and improves the detection performance because of its simplicity and locality preservation.

\begin{table}[h!]
\vspace{2pt}
    \footnotesize
    \centering
    \begin{tabular}{c|c|c}
        \hline
        Ordering Method &  NDS & mAP \\
        \hline
        No Ordering & 69.1& 63.0  \\
        \ours Ordering  & 70.2& 64.9 \\
        \hline
    \end{tabular}
    \vspace{-5pt}
    \caption{Effect of our proposed ordering for serializing the voxels into 1D sequence. We use 2-3-3 BEV backbone configuration in this experiment. We don't use any order shuffle or shift.}
    \vspace{-5pt}
    \label{tab:ordering}
\end{table}

\textbf{Grouping and Group Sizes.}
The implicit grouping provides two benefits: 1) improving inference speed and 2) encouraging a local-to-global learning. It can be seen in Table~\ref{tab:group} that by using our proposed implicit grouping with increasing group sizes, we improve both the accuracy and latency as compared to the baseline that has no grouping. Grouping with constant group sizes provides a very slight detection accuracy improvement. On the other hand, using decreasing group sizes is detrimental to detection performance as it conflicts with the growing effective receptive field as the inference goes deeper through the network. 

% The group partitioning technique enhances parallelization, thus accelerating the network. Our experiments also indicate a slight improvement in detection accuracy. In Table~\ref{tab:group}, we compare various group partitioning strategies, including a baseline without group partitioning, as well as configurations with constant, decreasing, and increasing group sizes. Specifically, the "Constant" setting uses a fixed group size of 256 across all \ourop blocks. The "Decreased" configuration indicates the group size is reduced from 1024 to 128 with (1024, 1024, 512, 512, 256, 256, 128, 128). The "Increased" configuration refers to grow the group size from 128 to 1024 with inverse order. The experiment results demonstrate that the increased group size not only achieves the highest detection accuracy. This improvement in accuracy is attributed to the increased group size facilitating gradual feature merging and enhanced voxel interaction while preserving parallelization.

\begin{table}[h!]
    \vspace{2pt}
    \footnotesize
    \centering
    \begin{tabular}{c|c|c}
        \hline
        Grouping Option &  NDS & mAP \\
        \hline
        No Grouping & 69.8& 64.8  \\
        Constant  & 69.9& 64.9 \\
        Decreasing & 69.7& 64.9 \\
        Increasing & 70.2& 65.9 \\
        \hline
    \end{tabular}
    \vspace{-5pt}
    \caption{Comparison of different grouping options. We use 2-3-3 BEV backbone configuration in this experiment. The grouping can significantly reduce the latency as table~\ref{tab:dim_bal}}
    \vspace{-5pt}
    \label{tab:group}
\end{table}

\textbf{\ourop.}
Our proposed \ourop serves as the core of \ours and provides an efficient, alternative way to process the features as compared to self-attention or 2D/3D sparse convolution. In order to better understand its effectiveness, we use DSVT~\cite{wang2023dsvt} as the baseline and replace the transformer layers with our \ourop layers. This is the only modification on DSVT in this experiment and we keep the other operations the same as the original DSVT, \eg, window shifting, alternating the voxel serialization between x-order and y-order, etc.

\begin{table}[h!]
    \vspace{2pt}
    \footnotesize
    \centering
    \begin{tabular}{l|c|c|c|c}
        \hline
        Baseline &Key Operation & NDS & mAP & FLOPs (G)  \\ 
        \hline
        \multirow{2}{*}{DSVT~\cite{wang2023dsvt}} & Transformer & 71.1& 66.4& 14.9\\
        & \ourop &71.2 &66.6 & 10.7\\ 
        \hline
    \end{tabular}
    \vspace{-5pt}
    \caption{Comparison of transformer and our proposed \ourop for processing voxels. We use DSVT~\cite{wang2023dsvt} and compare the performance of these two operations.}
    \vspace{-5pt}
    \label{tab:3dbackbone}
\end{table}

Table~\ref{tab:3dbackbone} shows the result. By replacing the transformer operations with our proposed \ourop, we can even slightly outperform the original DSVT. Moreover, this also reduces the computational cost (in terms of FLOPs) by more than 28\% in the 3D backbone. Note that \ourop will be even more efficient when the input 3D point cloud is large and/or the resolution is large, as its complexity scales linearly w.r.t. the number of input tokens, whereas self-attention scales quadratically.

\textbf{Depth of BEV Backbone.}
We use 3 ResNet configurations in the BEV backbone, which are 2-3-3, 3-3-4, and 4-5-5; the ResNet contains 3 stages with different number of layers in each stage. As shown in Table~\ref{tab:2dbackbone}, using more layers improves the detection performance at the cost of higher latency.

% The 2D backbone is scaled-up with different number of ResNet blocks in different stage as mentioned in Section~\ref{sec:2dbackbone}. There are three stage/resolution in 2D backbone. ResNet233 indicates 2 ResNet blocks in the first stage, 3 in the second, and 3 in the third. According to the experiment results in Table ~\ref{tab:2dbackbone}, increasing the depth of ResNet enhances detection accuracy, although this improvement eventually plateaus.
% \begin{table}[h!]
%     \footnotesize
%     \centering
%     \begin{tabular}{c|c|c|c|c|c}
%         \hline
%         \multirow{2}{*}{\#Layers} & \multirow{2}{*}{NDS} & \multirow{2}{*}{mAP} & \multirow{2}{*}{GMACs} & \multicolumn{2}{c}{Latency (ms)} \\
%         \cline{5-6}
%         & & & & NPU & GPU \\
%         \hline
%         8 & 70.2 & 64.9 & 157 & 33.8 & 16.6 \\
%         11 & 70.6 & 65.8 & 215 & 37.8 & 18.4 \\
%         14 & 70.8 & 65.7 & 272 & 41.7 & 21.2 \\
%         \hline
%     \end{tabular}
%     \vspace{-5pt}
%     \caption{Latency comparison of different scaled-up 2D backbones with detection accuracy and latency on device of NPU and GPU. There are 3 stage for the 2D backbone. L233 means 2 ResNet blocks in the first stage, 3 in the second, and 3 in the third.}
%     \label{tab:2dbackbone}
% \end{table}

\begin{table}[ht]
    \footnotesize
    \centering
    \begin{tabular}{c|c|c|c|c}
        \hline
        \multirow{2}{*}{ResNet Config.} & \multirow{2}{*}{NDS} & \multirow{2}{*}{mAP} & \multicolumn{2}{c}{Latency (ms)} \\
        \cline{4-5}
        & & & Hexagon NPU & Orin GPU \\
        \hline
        2-3-3 & 70.2 & 64.9 & 33.8 & 16.6 \\
        3-4-4 & 70.6 & 65.8 & 37.8 & 18.4 \\
        4-5-5 & 70.8 & 65.7 & 41.7 & 21.2 \\
        \hline
    \end{tabular}
    \vspace{-5pt}
    \caption{Effect of using different numbers of ResNet layers in BEV backbone.}
    \vspace{-5pt}
    \label{tab:2dbackbone}
\end{table}
\textbf{Large Kernel Size.} Reducing the kernel sizes from 11, 9, 7 to 5 leads to a drop in NDS from 70.8 to 70.2 on the nuScenes validation set. This indicates that larger kernels are beneficial for performance.
\vspace{-5pt}
\section{Conclusions and Limitation}
In this paper, we present \ours, a systematic and efficient solution for real-time LiDAR-based 3D object detection on resource-constrained platforms. By introducing \ourop, which leverages large-kernel convolutions, linear layers, and Hadamard products, along with hardware-friendly designs like one-time voxel serialization and implicit grouping, we eliminate costly operations such as sparse convolutions and transformers. \ours achieves up to 9.8× faster inference on embedded devices like Jetson Orin and Hexagon NPU, while maintaining competitive detection accuracy despite the inherent limitations of lightweight design.

Since we focus on efficiency for edge devices, using more memory- and compute-efficient operations can lead to some accuracy degradation, especially for small or sparse objects such as pedestrians and cyclists. Despite removing commonly used 3D sparse convolutions and transformer modules, our method achieves accuracy comparable to state-of-the-art approaches.
{
    \small
    \bibliographystyle{ieeenat_fullname}
    \bibliography{main}
}

% WARNING: do not forget to delete the supplementary pages from your submission 
% \input{sec/X_suppl}

\end{document}